\pdfoutput=1

\documentclass[11pt]{article}

\usepackage[final]{acl}

\usepackage{times}
\usepackage{latexsym}

\usepackage[T1]{fontenc}

\usepackage[utf8]{inputenc}

\usepackage{microtype}

\usepackage{inconsolata}

\usepackage{graphicx}

\usepackage{multirow}
\usepackage{booktabs}
\usepackage{tcolorbox}

\usepackage{xcolor}

\newcommand{\zhuowan}[1]{{#1}}

\newcommand{\ours}{\textsc{Self-Route}\xspace}
\newcommand{\longbench}{LongBench\xspace}
\newcommand{\infibench}{$\infty$Bench\xspace}
\newcommand{\gemini}{Gemini-1.5-Pro\xspace}
\newcommand{\gptthree}{GPT-3.5-Turbo\xspace}
\newcommand{\gptfouro}{GPT-4O\xspace}

\usepackage[capitalize]{cleveref}
\crefname{section}{Sec.}{Secs.}
\Crefname{section}{Section}{Sections}
\Crefname{table}{Table}{Tables}
\crefname{table}{Tab.}{Tabs.}
\Crefname{figure}{Figure}{Figures}
\crefname{figure}{Fig.}{Figs.}

\usepackage{xspace}
\makeatletter
\DeclareRobustCommand\onedot{\futurelet\@let@token\@onedot}
\def\@onedot{\ifx\@let@token.\else.\null\fi\xspace}

\def\eg{\emph{e.g}\onedot} 
\def\ie{\emph{i.e}\onedot}

\def\wrt{w.r.t\onedot} 

\makeatother

\title{Retrieval Augmented Generation or Long-Context LLMs?\\ A Comprehensive Study and Hybrid Approach}

\author{Zhuowan Li$^{1}$ \quad Cheng Li$^{1}$ \quad Mingyang Zhang$^{1}$ \\
{\bf Qiaozhu Mei$^{2}$\thanks{Visiting researcher to Google DeepMind.} \quad Michael Bendersky$^{1}$}\\
$^{1}$ Google DeepMind \quad $^{2}$ University of Michigan \\
{\tt  $^{1}$ \{zhuowan,chgli,mingyang,bemike\}@google.com \quad $^{2}$ qmei@umich.edu}
}

\begin{document}
\maketitle

\begin{abstract}

\textit{Retrieval Augmented Generation (RAG)} has been a powerful tool for \textit{Large Language Models (LLMs)} to efficiently process overly lengthy contexts. However, recent LLMs like Gemini-1.5 and GPT-4 show exceptional capabilities to understand long contexts directly. We conduct a comprehensive comparison between RAG and long-context \textit{(LC)} LLMs, aiming to leverage the strengths of both. We benchmark RAG and LC across various public datasets using three latest LLMs. Results reveal that when resourced sufficiently, LC consistently outperforms RAG in terms of average performance. However, RAG's significantly lower cost remains a distinct advantage. Based on this observation, we propose \ours, a simple yet effective method that routes queries to RAG or LC based on model self-reflection. \ours significantly reduces the computation cost while maintaining a comparable performance to LC. Our findings provide a guideline for long-context applications of LLMs using RAG and LC. 

\end{abstract}

\section{Introduction}

\textit{Retrieval augmented generation (RAG)} has been shown to be a both effective and efficient approach for large language models (LLMs) to leverage external knowledge. RAG retrieves relevant information based on the query and then prompts an LLM to generate a response in the context of the retrieved information. This approach significantly expands LLM's access to vast amounts of information at a minimal cost.

However, recent LLMs like Gemini and GPT-4 have demonstrated exceptional capabilities in understanding long contexts directly. For example, Gemini 1.5 can process up to 1 million tokens \cite{reid2024gemini}. 
\zhuowan{This prompts the need for a systematic comparison between long-context (LC) LLMs and RAG: on one hand, RAG conceptually acts as a prior, regularizing the attention of LLMs onto retrieved segments, thus avoiding the distraction of the irrelevant information and saving unnecessary attention computations; on the other hand, large-scale pretraining may enable LLMs to develop even stronger long-context capabilities. Therefore, we are motivated to compare RAG and LC, evaluating both their performance and efficiency.}

\begin{figure}[t!]
\begin{center}
\includegraphics[width=0.99 \linewidth]{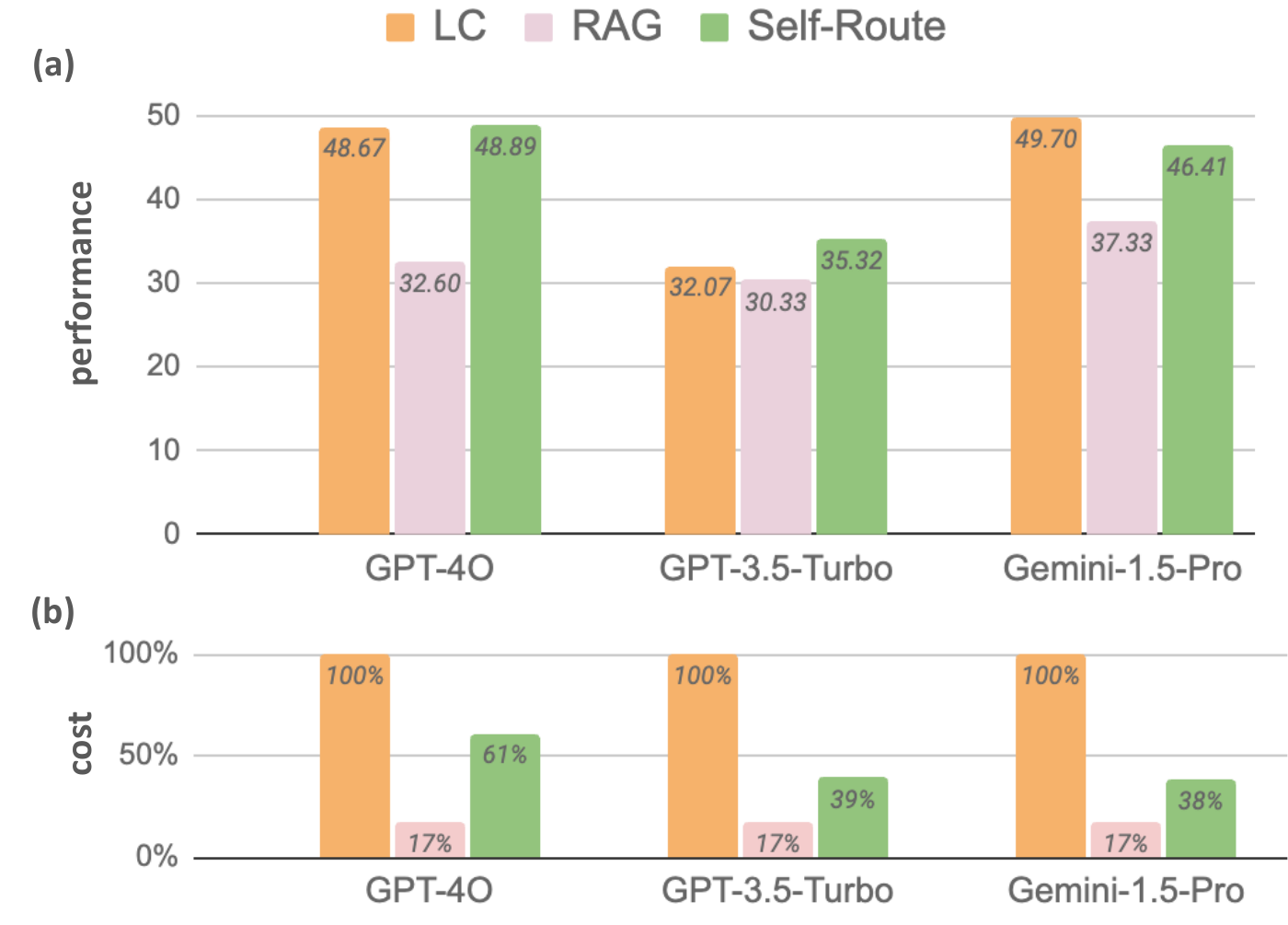}
\caption{While long-context LLMs (LC) surpass RAG in long-context understanding, RAG is significantly more cost-efficient. Our approach, \ours, combining RAG and LC, achieves comparable performance to LC at a much lower cost.}
\vspace{-1em}
\label{fig:intro}
\end{center}
\end{figure}

In this work, we systematically benchmark RAG and LC on various public datasets, gaining a comprehensive understanding of their pros and cons, and ultimately combining them to get the best of both worlds. 
Different from findings in previous work \cite{xu2023retrieval}, we find that LC consistently outperform RAG in almost all settings (when resourced sufficiently). This demonstrates the superior progress of recent LLMs in long-context understanding. 

\zhuowan{Despite the suboptimal performance, RAG remains relevant due to its significantly lower computational cost. In contrast to LC, RAG significantly decreases the input length to LLMs, leading to reduced costs, as LLM API pricing is typically based on the number of input tokens.~\cite{gemini-price, openai-price}\footnote{While retrieval may introduce extra cost, retrieval system is much easier to set up and can be hosted on customer side.}. Moreover, our analysis reveals that the predictions from LC and RAG are identical for over 60\% of queries. For these queries, RAG can reduce cost without sacrificing performance.}

Based on this observation, we propose \ours, a simple yet effective method that routes various queries to RAG or LC based on model self-reflection. With \ours, we significantly reduce the cost while achieving overall performance comparable to LC. For example, the cost is reduced by 65\% for \gemini and 39\% for \gptfouro. 

\cref{fig:intro} shows the comparisons of LC, RAG and \ours using three recent LLMs: \gptfouro, \gptthree and \gemini. In addition to quantitative evaluation, we provide a comprehensive analysis comparing RAG and LC, including common failure patterns of RAG, the trade-offs between cost and performance, and the results on additional synthetic datasets. Our analysis serves as a starting point, inspiring future improvements of RAG, and as a empirical guide for building long-context applications using RAG and LC.






\section{Related Work}

\noindent
\textbf{Long-context LLMs.}
There has long been efforts for enabling LLMs to handle long contexts \cite{guo2022longt5, beltagy2020longformer, chen2023extending}. While recent LLMs like Gemini-1.5 \cite{reid2024gemini}, GPT-4 \cite{achiam2023gpt}, Claude-3 \cite{claude35} achieve significantly larger context window size, long-context prompting is still expensive due to the quadratic computation cost of transformers regarding to the input token numbers. Recent work proposes methods to reduce cost by prompt compression \cite{jiang2023longllmlingua}, model distillation \cite{hsieh2023distilling}, or LLM cascading \cite{chen2023frugalgpt}.



\noindent
\textbf{Retrieval-augmented generation.}
Augmenting LLMs with relevant information retrieved from various sources \cite{lewis2020retrieval} has been successful in complementing LLMs with external knowledge. RAG achieves good performance on tasks like language modeling \cite{khandelwal2019generalization, shi2023replug} and QA \cite{guu2020retrieval, izacard2020leveraging}, with a significantly lower computation cost \cite{borgeaud2022improving}. Related to but different from our work, recently works augment RAG with correction \cite{yan2024corrective}, critique \cite{asai2023self}, verification \cite{li2023llatrieval}, or adaptive search \cite{wang2023self, cheng2024unified, jeong2024adaptive} to improve retrieval quality on knowledge-intensive tasks.



\noindent
\textbf{Long-context evaluation.}
Evaluating long-context models is challenging due to the difficulty in collecting and analyzing long texts. Recent researchers propose both synthetic tests like needle-in-a-haystack \cite{greg2023needle}, Ruler \cite{hsieh2024ruler}, or Counting Stars \cite{song2024counting}, and real datasets including \longbench \cite{bai2023longbench}, \infibench \cite{zhang2024infty}, L-Eval \cite{an2023eval}, and others \cite{shaham2022scrolls, yuan2024lv, maharana2024evaluating}. Evaluating on these datasets, recent works study the performance degradation over various context lengths \cite{levy2024same, hsieh2024ruler}, the lost-in-the-middle phenomenon \cite{liu2024lost}, and explore solutions \cite{kuratov2024search}. Related to our work, \citet{xu2023retrieval} compare RAG and long-context prompting and find that long-context models still lags behind RAG. This is different from \textbf{our} findings, possibly due to consideration of stronger LLMs and longer contexts in our work.



\section{Benchmarking RAG versus LC}
\label{sec:benchmark}

\subsection{Datasets and metrics}


\zhuowan{We evaluate on a subset of datasets from \longbench \cite{bai2023longbench} and \infibench \cite{zhang2024infty}, which are recent benchmarks containing a collection of new and existing datasets for LLM evaluation, covering both synthetic and real texts in multiple languages. \longbench contains a collection of 21 datasets, with an average context length of 7k words. \infibench consists of even longer contexts with an average length of 100k tokens. 

Among the datasets, we mainly focus on tasks that are (a) in English, (b) real, and (c) query-based (\eg summarization tasks do not contain queries for retrieving relevant information). This results in 7 datasets from \longbench including NarrativeQA \cite{kovcisky2018narrativeqa}, Qasper \cite{dasigi2021dataset}, MultiFieldQA \cite{bai2023longbench}, HotpotQA \cite{yang2018hotpotqa}, 2WikiMultihopQA \cite{ho2020constructing}, MuSiQue \cite{trivedi2022musique}, QMSum \cite{zhong2021qmsum}; and 2 datasets from \infibench including En.QA and EN.MC. Please refer to \cref{sec:dataset} for more details.
Additionally, in \cref{sec:synthetic}, we will provide an ablation a synthetic datasets PassKey from \infibench.}

For evaluation metrics, we report F1 scores for the open-ended QA tasks, accuracy for the multi-choice QA tasks, and ROUGE score for the summarization tasks. 

\subsection{Models and Retrievers}

Three latest LLMs are evaluated, including \gemini \cite{reid2024gemini}, \gptfouro \cite{gpt4o}, and \gptthree \cite{gpt3.5} \footnote{gpt-3.5-turbo-0125, gpt-4o-2024-05-13}. \gemini is a recent long-context LLM from Google, supporting up to 1 million tokens. \gptfouro, the newest lightweight yet strong LLM from OpenAI, supports 128k tokens. \gptthree supports 16k tokens. 

Two retrievers are used in our study: Contriever \cite{izacard2021unsupervised}, which is a contrastively trained dense retriever outperforming BM25 on BEIR datasets, and Dragon \cite{lin2023train}, which is a recent generalizable dense retriever achieving high performance in both supervised and zero-shot settings without complex late interaction. Following \cite{xu2023retrieval}, we divide long contexts into chunks of 300 words, and select the top $k$ chunks (default $k=5$) based on the cosine similarity of the query embedding and the chunk embeddings. The chunks are ordered by the similarity scores, with the chunk index prepended at the beginning.

\zhuowan{Since black-box LLMs are pretrained on unknown datasets, the leakage of evaluation datasets may occur. Especially, some of the evaluation datasets are based on Wikipedia, which has likely been seen by LLMs during during. In some cases, we find that model may predict the correct answer using exactly the same words as the groundtruth (\eg ``meticulously''), even when they do not appear in the provided context. In our experiment, we try mitigating this issue by prompting the model to answer \texttt{``based only on the provided passage''} for both RAG and LC. It remains an open question how to address the data leakage issue in LLM evaluation.}

\subsection{Benchmarking results}
We benchmark the performance of LC and RAG across the nine datasets, using three recent LLMs: \gemini, \gptfouro and \gptthree. 
\zhuowan{\cref{tab:benchmark} presents the results using the Contriever retriever, where rows *-1 and rows *-2 present the benchmarking results for LC and RAG respectively. Results using the Dragon retriever will be discussed in \cref{sec:retrievers} and \cref{tab:dragon}.}

As shown in \cref{tab:benchmark}, LC consistently outperforms RAG for all the three models, with a significant margin. On average, LC surpasses RAG by 7.6\% for \gemini, 13.1\% for \gptfouro, and 3.6\% for \gptthree. Noticeably, the performance gap is more significant for the more recent models (\gptfouro and \gemini) compared to \gptthree, highlighting the exceptional long-context understanding capacity of the latest LLMs.

However, there is an exception observed on the two longer datasets from \infibench (\ie, En.QA and En.MC), where RAG achieves higher performance than LC for \gptthree. This result deviates from the overall trend, likely due to the significantly longer context in these datasets (147k words on average) compared with the limited context window (16k) of \gptthree. This finding highlights the effectiveness of RAG when the input text considerably exceeds the model's context window size, emphasizing a specific use case of RAG.

\begin{table*}[ht!]
\begin{center}
\resizebox{1.0\linewidth}{!}{ 
\begin{tabular}{lll|r|rrrrrrrrr}
\toprule
 &  &  & \textbf{Avg} & \textbf{Narr} & \textbf{Qasp} & \textbf{Mult} & \textbf{Hotp} & \textbf{2Wiki} & \textbf{Musi} & \textbf{Sum} & \textbf{En.QA} & \textbf{En.MC} \\ \midrule
 & 1-1 & LC & \textbf{49.70} & \textbf{32.76} & \textbf{47.83} & \textbf{52.33} & \textbf{61.85} & \textbf{62.96} & 40.22 & \textbf{20.73} & \textbf{43.08} & \textbf{85.57} \\
\multirow{5}{*}{Gemini-1.5-Pro} & 1-2 & RAG & 37.33 & 22.54 & 44.68 & 49.53 & 48.36 & 54.24 & 26.56 & 19.51 & 19.46 & 51.09 \\
 & 1-3 & \ours & 46.41 & 28.32 & 45.23 & 51.47 & 55.18 & 62.68 & \textbf{40.66} & 19.77 & 37.51 & 76.86 \\ \cline{2-13}
 & 1-4 & answerable \% & 76.78 & 73.00 & 85.00 & 96.67 & 84.50 & 81.00 & 58.50 & 93.50 & 56.41 & 62.45 \\
 & 1-5 & token \% & 38.39 & 23.07 & 49.93 & 36.88 & 32.97 & 53.49 & 56.14 & 17.96 & 42.25 & 32.84 \\ \midrule
\multirow{5}{*}{GPT-4O} & 2-1 & LC & 48.67 & \textbf{32.78} & 44.54 & \textbf{55.28} & \textbf{62.42} & \textbf{70.69} & 41.65 & \textbf{21.92} & 32.36 & 76.42 \\
 & 2-2 & RAG & 32.60 & 18.05 & 46.02 & 50.74 & 36.86 & 50.21 & 16.09 & 19.97 & 14.43 & 41.05 \\
 & 2-3 & \ours & \textbf{48.89} & 31.36 & \textbf{47.99} & 53.17 & 62.14 & 70.14 & \textbf{41.69} & 21.31 & \textbf{34.95} & \textbf{77.29} \\ \cline{2-13}
 & 2-4 & answerable \% & 57.36 & 44.00 & 67.50 & 94.00 & 52.50 & 62.00 & 30.00 & 92.00 & 27.07 & 47.16 \\
 & 2-5 & token \% & 61.40 & 66.40 & 72.25 & 39.65 & 65.79 & 77.05 & 85.00 & 20.26 & 73.01 & 53.21 \\ \midrule
\multirow{5}{*}{GPT-3.5-Turbo} & 3-1 & LC & 32.07 & 23.34 & \textbf{42.96} & 49.19 & 45.33 & 41.04 & 17.92 & 19.61 & 14.73 & 34.50 \\
 & 3-2 & RAG & 30.33 & 18.22 & 38.15 & 49.21 & 37.84 & 35.16 & 16.41 & 18.94 & 15.39 & 43.67 \\
 & 3-3 & \ours & \textbf{35.32} & \textbf{24.06} & 38.65 & \textbf{52.07} & \textbf{47.28} & \textbf{44.62} & \textbf{34.44} & \textbf{19.88} & \textbf{22.03} & \textbf{44.54} \\ \cline{2-13}
 & 3-4 & answerable \% & 74.10 & 71.50 & 80.00 & 91.33 & 68.50 & 69.00 & 47.00 & 93.50 & 50.43 & 95.63 \\
 & 3-5 & token \% & 38.85 & 20.56 & 55.08 & 35.29 & 48.70 & 65.91 & 65.08 & 16.40 & 38.17 & 4.50 \\ \bottomrule
\end{tabular}
}
\end{center}
\caption{Results of \gemini, \gptthree, and \gptfouro using the Contriever retriever. LC consistently outperforms RAG, while \ours achieves performance comparable to LC using much less tokens.}
\vspace{-0.5em}
\label{tab:benchmark}
\end{table*}

\section{Self-Route}
\label{sec:route}

\subsection{Motivation}
\begin{figure}[h]
\begin{center}
    \includegraphics[width=0.95 \linewidth]{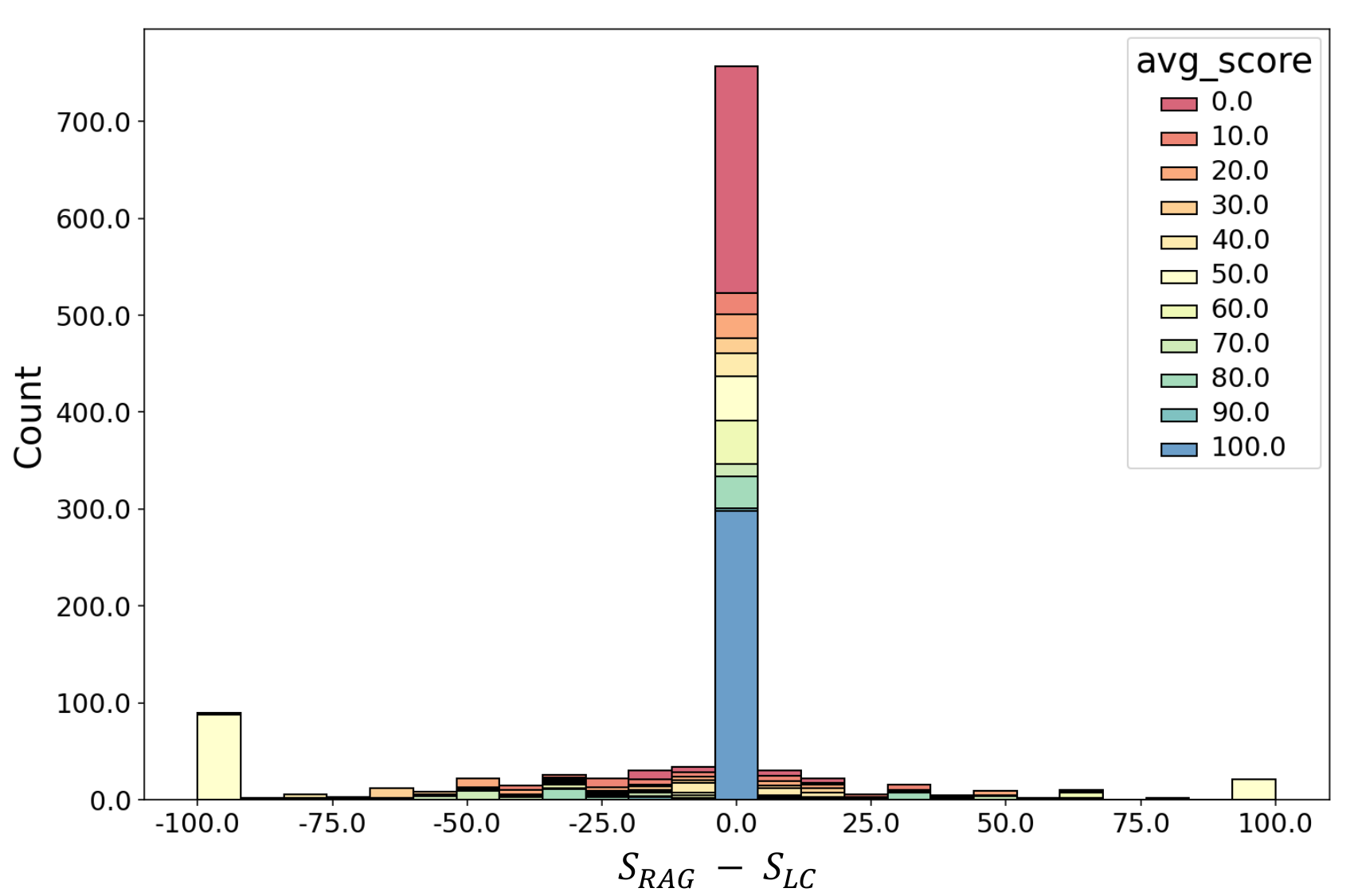}
    \caption{Distribution of the difference of prediction scores between RAG and LC (computed \wrt groundtruth labels). RAG and LC predictions are highly identical, for both correct and incorrect ones.}
    \label{fig:dist}
\end{center}
\end{figure}

As demonstrated in \cref{sec:benchmark}, RAG lags behind long-context LLMs in terms of performance. However, despite this performance gap, we surprisingly find a high degree of overlap in their predictions, as illustrated in \cref{fig:dist}. 

\cref{fig:dist} displays the distribution of the differences between RAG prediction scores $S_{RAG}$ and LC prediction scores $S_{LC}$, specifically $S_{RAG}-S_{LC}$ \zhuowan{(the scores are multiplied by 100 to be scaled to 1-100)}. These scores $S$ represent the evaluation of model predictions against the groundtruth. Notably, for most queries, RAG scores and LC scores are highly similar. In fact, for 63\% queries, the model predictions are exactly identical; and for 70\% queries, \zhuowan{the score difference is less than 10 (absolute value)}. Interestingly, the identical predictions are not necessarily correct, as shown by the varying colors representing the average score, \ie, $(S_{RAG}+S_{LC})/2$. This observation suggests that RAG and LC tend to make not only the same correct predictions but also similar errors.

This finding motivates us to leverage RAG for the majority of queries, reserving computationally more expensive LC for a small subset of queries where it truly excels. By doing so, RAG can significantly reduce computational costs without sacrificing overall performance.

\subsection{Self-Route}

Based on the above motivation, we propose \ours, a simple yet effective method combining RAG and LC to reduce cost while maintaining a performance comparable to LC. \ours utilizes LLM itself to route queries based on self-reflection, under the assumption that LLMs are well-calibrated in predicting whether a query is answerable given provided context.

Concretely, our method consists of two steps: a RAG-and-Route step and a long-context prediction step. In the first step, we provide the query and the retrieved chunks to the LLM, and prompt it to predict whether the query is answerable and, if so, generate the answer. This is similar to standard RAG, with one key difference: the LLM is given the option to decline answering with the prompt \texttt{``Write unanswerable if the query can not be answered based on the provided text''}. For the queries deemed answerable, we accept the RAG prediction as the final answer. For the queries deemed unanswerable, we proceed to the second step, providing the full context to the long-context LLMs to obtain the final prediction (\ie, LC). 

As our results will demonstrate, most queries can be solved by the first RAG-and-Route step (\eg, 82\% for \gemini), with only a small portion requiring the following long-context prediction step. Since the RAG-and-Route step only needs the retrieved chunks (\eg, 1.5k tokens) as input, which is significantly shorter than the full contexts (\eg, 10k - 100k tokens), the overall computation cost is substantially reduced. Detailed token count analysis will be provided in the results.

\subsection{Results}
Rows *-3 to *-5 in \cref{tab:benchmark} present the results of our method, utilizing the three LLMs. Rows *-3 report the performance. Rows *-4 show the percentage of answerable queries, as predicted in the RAG-and-Route step. Rows *-5 display the percentage of tokens used by our method, compared to that of LC. 
In terms of performance (rows *-3), \ours significantly outperforms RAG, achieving results comparable to LC. Across all three models, \ours surpasses RAG (rows *-2) by over 5\%. Compared to LC (rows *-1), there is a slight performance drop for \gptfouro (-0.2\%) and \gemini (-2.2\%), but an improvement for \gptthree (+1.7\%).

All three LLMs consistently route more than half of queries towards RAG, as shown in rows *-4. For \gemini, the answerable percentage even reaches 81.74\% (row 1-4). This indicates that RAG may answer most queries without the need for LC, confirming our initial motivation.

Due to the high answerable rate, the number of tokens required is significantly reduced (rows *-5). For example, \gptfouro uses only 61\% tokens while achieving comparable performance (46.83) with LC (47.04), Gemini-1.5-Pro uses 38.6\% of the tokens. Since the computation cost of the transformer-based LLMs is quadratic to token count, and most LLM APIs charge based on token count \cite{openai-price, gemini-price}, this lower token count translates to substantial cost savings.

\zhuowan{On longer datasets, the advantage of our method is more pronounced for OpenAI models, but less significant for Gemini. For instance, for \gptfouro, \ours outperforms LC by 2.3\% and 7.4\% respectively on EN.QA and EN.MC, which contain longer contexts. For \gptthree, the advantage margins are even larger. However, for \gemini, the performance is lower than LC. These different behaviors are possibly due to the difference in LLM alignments, \ie, OpenAI models are more likely to reject answering using RAG, leading to a lower answerable percentage but higher accuracy, which results in a different performance-cost trade-off compared with \gemini.}

\section{Analysis}

\subsection{Ablations of k}

\begin{figure}[h]
\begin{center}
    \includegraphics[width=0.95 \linewidth]{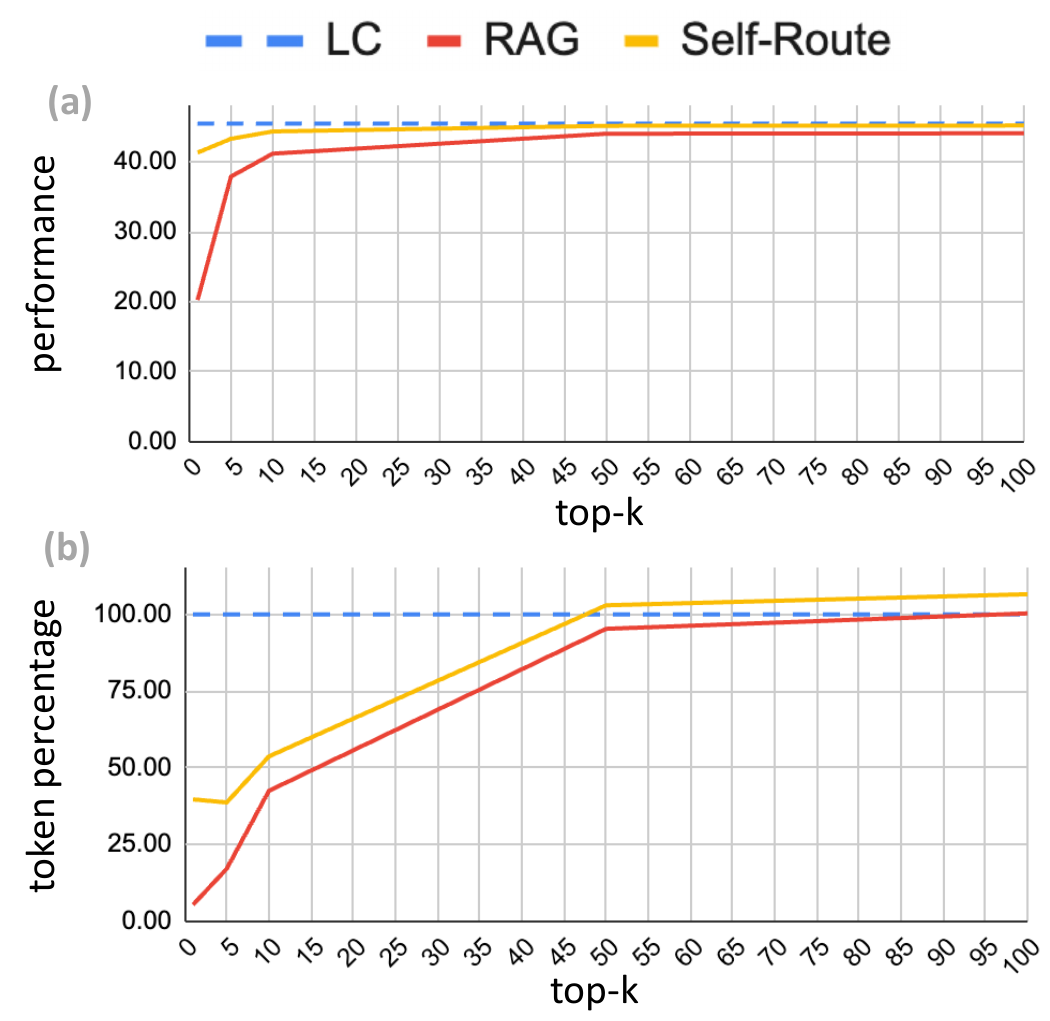}
    \caption{Trade-off curves between (a) model performance and (b) token percentage as a function of $k$.}
\vspace{-1em}
    \label{fig:curve}
\end{center}
\end{figure}

Both RAG and \ours relies on the top-$k$ retrieved text chunks. The larger $k$ is, the longer context are fed into LLMs for RAG prediction as well as routing, resulting in different costs versus performances. To study the influence of $k$, in \cref{fig:curve}, we plot the performance and cost (\ie input token percentage) curves when different $k$s are used.

In terms of performance, for both RAG and \ours, a larger $k$ leads to better performance. While $k$ increases, more and more chunks are fed into the LLMs, thus the performance gradually improves to approach LC. As can be seen in from the curves, the advantage of \ours is the most significant for smaller $k$. For example, when $k=1$, RAG gets from 20.24\% while \ours gets 37.9\%, while when $k$ is larger than 50, all three methods get similar performance.

However, the trend of cost is not monotonous for \ours. As seen, the cost reaches its minimum at $k=5$. This is because when $k$ increases, the cost of RAG (and routing) increases, but more queries are routed to RAG from LC, thus the overall cost may decrease. The sweet point of $k$ might be different for each dataset, \eg on average, $k=5$ has the lowest cost as shown in the curves, but on some datasets, especially ones that contain extractive questions which does not need multi-hop reasoning (like NarrativeQA and QMSum), $k=1$ leads to the lowest cost. This indicates that the optimal $k$ depends on the nature of the task, as well as the performance requirement. We encourage future researchers to look for different $k$s when applying our method to various applications.

\begin{table*}[t]
\begin{center}
\resizebox{0.95\linewidth}{!}{ 
\begin{tabular}{lll|r|rrrrrrrrr}
\toprule
 &  &  & \textbf{Avg} & \textbf{Narr} & \textbf{Qasp} & \textbf{Mult} & \textbf{Hotp} & \textbf{2Wiki} & \textbf{Musi} & \textbf{Sum} & \textbf{En.QA} & \textbf{En.MC} \\ \midrule
 & 1 & LC & 49.70 & 32.76 & 47.83 & 52.33 & 61.85 & 62.96 & 40.22 & 20.73 & 43.08 & 85.57 \\
\multirow{5}{*}{Dragon} & 2 & RAG & 38.09 & 21.91 & 44.33 & 53.08 & 51.61 & 50.05 & 30.47 & 19.93 & 21.25 &  50.22 \\
 & 3 & combine & 46.81 & 28.50 & 43.82 & 54.62 & 56.58 & 60.62 & 40.66 & 20.07 & 37.79 & 78.60 \\ \cline{2-13}
 & 4 & RAG ratio & 77.88 & 74.00 & 84.00 & 97.33 & 86.00 & 77.00 & 66.00 & 95.50 & 61.25 & 59.83 \\
 & 5 & Token ratio & 37.87 & 19.31 & 54.15 & 34.78 & 32.64 & 55.65 & 48.16 & 16.64 & 38.71 & 40.83 \\ \bottomrule
\end{tabular}
}
\end{center}
\vspace{-0.5em}
\caption{Results for Gemini-1.5-Pro using Dragon retriever.}
\vspace{-0.5em}
\label{tab:dragon}
\end{table*}

\subsection{Why does RAG fail?}
\label{sec:failure}

To gain a better understanding of why RAG lags behind LC, we analyze the failure reasons for the examples that cannot be answered by RAG. We first manually check some examples for which our RAG-and-Route step predicts ``unanswerable'' and summarize four typical failure reasons, then prompt LLM to classify all the examples.

The four reasons include: (A) The query requires multi-step reasoning so the results of previous steps are needed to retrieve information for later steps, \eg \texttt{``What nationality is the performer of song XXX''}. (B) The query is general, \eg \texttt{``What does the group think about XXX''}, which is challenging for the retriever to formulate a good query. 
(C) The query is long and complex, which is challenging for the retriever to understand. However, answering this kind of questions is arguably, an advantage of LLMs. (D) The query is implicit, demanding a thorough understanding of the entire context. For instance, in a lengthy conversational narrative about a space voyage, a question like \texttt{``What caused the shadow behind the spaceship?''} requires readers to connect the dots and deduce the answer, as there is no explicit mention of the shadow when the cause is revealed.

\begin{figure}[h]
\begin{center}
    \includegraphics[width=1.0 \linewidth]{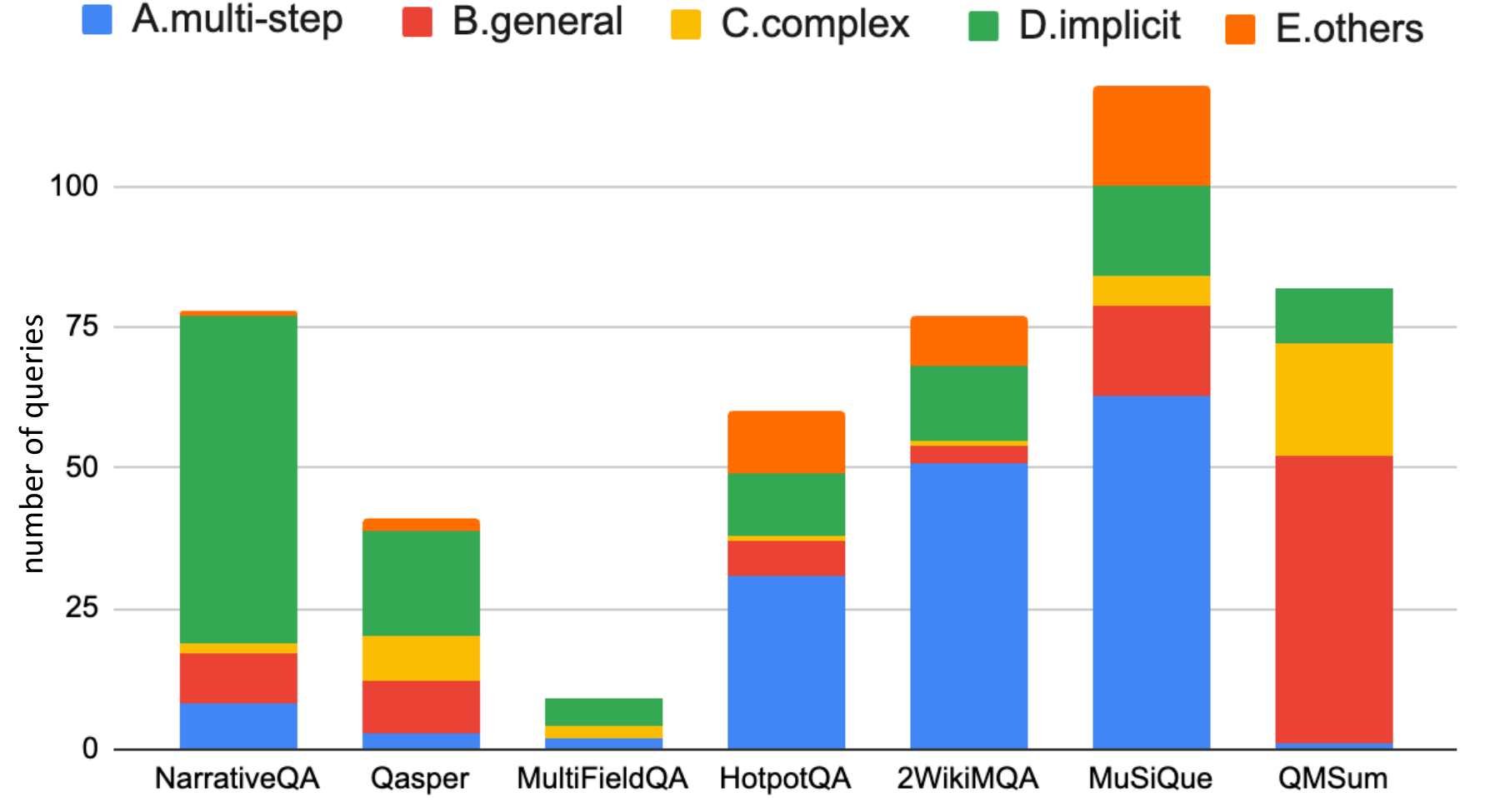}
\vspace{-1em}
    \caption{Distribution of typical RAG failure reasons.}
\vspace{-0.5em}
    \label{fig:failure}
\end{center}
\end{figure}

\zhuowan{Using these reasons, we prompt \gemini with few-shot in-context examples that we manually annotated, to classify all the unanswerable examples into these four categories, plus an ``other" option.} \cref{fig:failure} shows the distribution of failure reasons on the seven datasets in \longbench. Each dataset may contain different number of RAG failure cases, resulting in various bar heights. The distribution patterns are consistent with the nature of the datasets. For example, the three Wikipedia-based multi-hop reasoning datasets (HotpotQA, 2WikiMQA, MuSiQue) are challenging for RAG because of multi-step retrieval as shown in blue. For NarrativeQA, which are long stories containing a lot of dialogues, most failure cases are due to implicit queries that requires understanding the whole context (shown in green). For QMSum, which is a summarization dataset contains open-ended questions, failures are mostly due to general queries (shown in red). \zhuowan{We manually checked the examples classified as ``others'' and find that most of them are actually multi-step questions, often with ambiguities, which poses challenges for answering.}

We hope this failure analysis inspires future improvements of RAG. \zhuowan{For example, engaging chain-of-thought \cite{wei2022chain} into RAG may help address the multi-step questions, and revisiting query understanding techniques like query expansion \cite{lv2009adaptive, zhai2001model} may help with the general queries and complex queries. We are also glad to see recent efforts towards the direction \cite{chan2024rq, ma2023query}.}

\subsection{Different retrievers}
\label{sec:retrievers}

The results using a retriever, Dragon, is shown in \cref{tab:dragon} based on \gemini. As can be seen, the results are consistent with Contriever, for all of LC, RAG, and \ours, showing that our findings are generalizable across retrievers.


\subsection{Results on synthetic data}
\label{sec:synthetic}

In this study, we mainly focus on real datasets, with a consideration that results on synthetic data, which are artificially created by researchers, may subject to dataset artifacts. \zhuowan{We notice some methods that researchers adopted to create synthetic long context datasets may unconsciously, but largely, influence the performance comparison between RAG and LC.} For example, here we describe the results on the ``PassKey'' dataset in \infibench and its variations. 

This ``PassKey'' dataset presents a needle-in-a-haystack test, where a sentence with a passkey (\eg \texttt{``the passkey is 123456''}) is hidden within chunks of irrelevant text, and the model is asked to answer the question \texttt{``What is the passkey''}. The task requires strong retrieval capability. On this dataset, RAG achieves 80.34\% accuracy, outperforming LC, which gets 65.25\% using \gemini. However, if the query is slightly modified as \texttt{``What is the special token hidden inside the texts''}, RAG accuracy sharply drops to only 4.58\%, while LC keeps roughly the same (69.32\%). Another example: if the chunks contain two passkeys and the query is \texttt{``Which passkey is larger? First or second?''}, then RAG (47.63\%) under-performs LC (64.24\%) as well. \cref{tab:synthetic} summarizes the results, which demonstrates that the evaluation highly subjects to artifacts in dataset construction, showing limitation of synthetic testing.

\begin{table}[]
\begin{center}
\resizebox{0.85\linewidth}{!}{ 
\begin{tabular}{lrr}
\toprule
 & \textbf{RAG} & \textbf{LC} \\ \midrule
Original & 80.34 & 65.25 \\ \midrule
Variant-1: ``special token'' & 4.58 & 69.32 \\ 
Variant-2: ``which is larger'' & 47.63 & 64.24 \\ \bottomrule
\end{tabular}
}
\end{center}
\vspace{-1em}
\caption{Synthetic dataset may unconsciously contain artifacts that influence the comparison results.}
\vspace{-0.5em}
\label{tab:synthetic}
\end{table}


\subsection{Exclusion of LLM's internal knowledge}
Ideally, the comparison in this paper should exclude the model's internal knowledge (\ie, parametric knowledge) so that the model's performance are solely based on its capability to understand long contexts. In our study, this internal knowledge is excluded by utilizing the prompt ``based only on the provided passage'', which we empirically find is a simple yet effective method. Here we discuss the effectiveness of this method, as well as alternative methods to exclude external knowledge.

First, we validate the effectiveness of the simple prompt ``based only on the provided passage''. \cref{tab:based_only_on} compares the performance (long-context) of \gemini with and without this prompt. As shown, using this prompt consistently limits the model’s performance (average performance drops from 50.57 to 45.53), which indicates that using this simple instruction can already effectively limit the usage of the model’s parametric knowledge.

\begin{table}[h]
\begin{center}
\resizebox{1.0\linewidth}{!}{ 
\begin{tabular}{l|cc}
\toprule
 & \begin{tabular}[c]{@{}c@{}}without\\ "based only on ..."\end{tabular} & \begin{tabular}[c]{@{}c@{}}with \\ "based only on ..."\end{tabular} \\ \midrule
NarrativeQA & 36.35 & 32.76 \\
Qasper & 50.69 & 47.83 \\
MultiFieldQA & 56.07 & 52.33 \\
HotpotQA & 66.47 & 61.85 \\
2WikiMQA & 68.97 & 62.96 \\
Musique & 54.56 & 40.22 \\
QMSum & 20.87 & 20.73 \\
En.QA & 49.20 & 43.08 \\
En.MC & 90.83 & 85.57 \\ \midrule
\textbf{Avg} & \textbf{50.57} & \textbf{45.53} \\ \bottomrule
\end{tabular}
}
\end{center}
\caption{Comparison of the long-context performance of \gemini, using the prompt with and without ``based only on the provided passage''.}
\label{tab:based_only_on}
\end{table}

Second, as an alternative method to exclude internal knowledge, we remove the questions where the model can correctly answer without any contexts (\ie, commonsense questions), and report the model's performance only on the non-commonsense questions. \cref{tab:exclude_commonsense} shows the performance of \gemini and \gptthree on all the questions from the MuSiQue dataset, as well as their performance on the non-commonsense subset\footnote{Different models may learn different internal knowledge, resulting in different numbers of non-commonsense questions. For example, \gptthree gets 14.53 performance on MuSiQue while \gemini gets 23.58 using only internal knowledge,}. As shown, after excluding the commonsense questions, the trend remains the same. 

\begin{table}[h]
\begin{center}
\resizebox{1.0\linewidth}{!}{ 
\begin{tabular}{l|cc|cc}
\toprule
 & \multicolumn{2}{c|}{\textbf{all questions}} & \multicolumn{2}{c}{\textbf{w/o commonsense}} \\
 & {Gemini} & {GPT-3.5} & {Gemini} & {GPT-3.5} \\ \midrule
\# questions & 200 & 200 & 133 & 150 \\ \midrule
LC & 40.22 & 17.92 & \textbf{31.76} & 13.00 \\
RAG & 26.56 & 16.41 & 15.51 & 13.05 \\
Self-Route & \textbf{40.66} & \textbf{34.44} & 31.32 & \textbf{19.76} \\ \midrule
answerable \% & 58.50 & 47.00 & 52.63 & 45.33 \\
token \% & 56.14 & 65.08 & 48.46 & 53.43 \\ \bottomrule
\end{tabular}
}
\end{center}
\vspace{-0.5em}
\caption{Results on MuSiQue on all questions, and on the subset of non-commonsense questions (\ie, excluding questions that can be answered without contexts).}
\label{tab:exclude_commonsense}
\end{table}

That said, a more thorough study to explore various methods for controlling the usage of model's internal knowledge, and to study the source of internal knowledge (\eg LLM"s world knowledge or dataset leakage), will be valuable future work, which we hope can be further investigated.

\section{conclusion}

This paper presents a comprehensive comparison of RAG and LC, highlighting the trade-offs between performance and computational cost. While LC demonstrate superior performance in long-context understanding, RAG remains a viable option due to its lower cost and advantages when the input considerably exceeds the model’s context window size. Our proposed method, which dynamically routes queries based on model self-reflection, effectively combines the strengths of both RAG and LC, achieving comparable performance to LC at a significantly reduced cost. We believe our findings contribute valuable insights for the practical application of long-context LLMs and pave the way for future research in optimizing RAG techniques.

\subsection*{Acknowledgements}
\noindent
We would like to thank Weize Kong, Tao Chen, Jeffrey Dudek and Spurthi Amba Hombaiah for their helpful comments and suggestions, as well as the anonymous reviewers for the valuable discussions.

\bibliography{custom}


\onecolumn
\appendix

\section{Dataset details}
\label{sec:dataset}

We evaluate on 7 datasets from \longbench \cite{bai2023longbench}. \textbf{NarrativeQA} \cite{kovcisky2018narrativeqa} is a question answering dataset, where the context is a long story like a novel or a movie script. \textbf{Qasper} \cite{dasigi2021dataset} focuses on question answering over academic NLP papers and is annotated by NLP practitioners. \textbf{MultiFieldQA}, originally proposed in \longbench, contains human-annotated QA over documents and articles from multiple sources, including legal documents, government reports, encyclopedias, academic papers, etc. \textbf{HotpotQA} \cite{yang2018hotpotqa} contains two-hop questions written by native English speakers that requires reasoning over two related Wikipedia paragraphs in the long context. \textbf{2WikiMultihopQA} \cite{ho2020constructing} contains up to 5-hop questions that are synthesized through manually designed templates, ensuring that they cannot be solved through shortcuts. The questions in \textbf{MuSiQue} \cite{trivedi2022musique} are up to 4-hop, first constructed from single-hop question compositions, and then paraphrased by annotators for linguistic diversity. \textbf{QMSum} \cite{zhong2021qmsum} is a query-based summarization dataset over meeting scripts from multiple domains.

We evaluate on 2 datasets from \infibench \cite{zhang2024infty}. \textbf{En.QA} contains human-annotated question-answer pairs for long novels, with key entity names manually replaced in order to avoid knowledge leakage due to model pretraining. \textbf{EN.MC} is annotated similarly to En.QA, but differs in that the model is presented with four challenging answer choices written by the annotators.

\cref{tab:dataset} shows the details of the datasets, including the number of queries in each evaluation dataset and the average context length (\ie number of words).

\begin{table*}[h]
\begin{center}
\begin{tabular}{clcc}
\toprule
 & \multicolumn{1}{l}{} & \multicolumn{1}{l}{\textbf{Num. Query}} & \multicolumn{1}{l}{\textbf{Avg. Length}} \\ \midrule
\multirow{7}{*}{\begin{tabular}[c]{@{}c@{}}LongBench \\ \cite{bai2023longbench}\end{tabular}} & \textbf{NarrativeQA} & 200 & 18,395 \\
 & \textbf{Qasper} & 200 & 3,599  \\
 & \textbf{MultiFieldQA} & 150 & 4,539  \\
 & \textbf{HotpotQA} & 200 & 9,133  \\
 & \textbf{2WikiMultihopQA} & 200 & 4,873  \\
 & \textbf{MuSiQue} & 200 & 11,196  \\
 & \textbf{QMSum} & 200 & 10,533  \\ \midrule
\multirow{2}{*}{\begin{tabular}[c]{@{}c@{}}\infibench \\ \cite{zhang2024infty}\end{tabular}} & \textbf{En.QA} & 351 & 150,374  \\
 & \textbf{En.MC} & 229 & 142,622 \\
\bottomrule
\end{tabular}
\end{center}
\caption{Dataset statistics.}
\label{tab:dataset}
\end{table*}

\section{Ablations of k}
\cref{tab:curve} shows the performance and token ratio for different $k$, which corresponds to \cref{fig:curve}. The performance of LC, which serves as an upper bound, is 45.53. The token ratio is computed the token counts for RAG or \ours divided the number of tokens required by LC.


\begin{table*}[h]
\begin{center}
\begin{tabular}{r|cc|cc} \toprule
\textbf{} & \multicolumn{2}{c|}{\textbf{performance}} & \multicolumn{2}{c}{\textbf{token ratio}} \\
\textbf{top-k} & \textbf{RAG} & \textbf{Self-Route} & \textbf{RAG} & \textbf{Self-Route} \\ \midrule
\textbf{1} & 20.24 & 41.35 & 5.26 & 39.64 \\
\textbf{5} & 37.92 & 43.33 & 17.02 & 38.63 \\
\textbf{10} & 41.20 & 44.38 & 42.42 & 53.66 \\
\textbf{50} & 44.06 & 45.19 & 95.29 & 102.97 \\
\textbf{100} & 44.12 & 45.23 & 100.32 & 106.59 \\ \bottomrule
\end{tabular}
\end{center}
\caption{Performance and token ratio for different $k$. This table corresponds to \cref{fig:curve}.}
\label{tab:curve}
\end{table*}

\section{Prompts}
\label{sec:prompts}

\cref{tab:prompt} shows the prompts for each dataset in our study. The prompts are modified from the released prompts as in \longbench \cite{bai2023longbench} and \infibench \cite{zhang2024infty}. \cref{tab:prompt_failure} shows the prompts used in the failure case study as in \cref{sec:failure}.

\begin{table*}[h!]\centering
\begin{minipage}{0.99\columnwidth}\vspace{0mm}
\begin{tcolorbox} 
\centering
\small
\hspace{-6mm}
\begin{tabular}{p{0.99\columnwidth}}
\begin{minipage}{0.99\columnwidth}\vspace{0mm}
You are given some text chunks from an article, and a question. The text chunks are retrieved by an external retriever. Now: \\

(1) Tell whether the question can be answered based only on the provided text chunks.\\
(2) If the question can be answered, answer the question based on the texts as concisely as you can, using a single phrase if possible. \\
(3) If the question cannot be answered, choose the reason from the following: \\

A. The question needs multistep reasoning, thus it is hard to retrieve all the relevant chunks. For example, "What nationality is the performer of song You Can?" contains two steps: find the performer, then find the nationality of the performer. Other examples include "Where does the director of film Wine Of Morning work at?", "What is another notable work made by the author of Miss Sara Sampson?"

B. The question is a general query, thus it is hard to retrieve relevant chunks. For example, "What did the group think about Dave leaving?" is general because the group may include multiple persons, and they can have different thinkings.

C. The question is long and complex, which is hard for the retriever to encode it to retrieve relevant chunks. For example, "What did Julie Morgan elaborate on the online survey when talking about the evaluations on the legitimacy of the children's rights, protection and demands?", "The Huskies football team were invited to the Alamo Bowl where they were defeated by a team coached by Art Briles and who played their home games at what stadium?"

D. The question is not explicit and requires comprehensive understanding of the whole story and cannot be solved using retrieval-augmented generation. For example, "What caused the shadow behind Koerber's ship?" needs a comprehensive understanding of the whole story. Another example like "How many words are there in the article" also requires the complete article.

E. Others.

Keep the above reasons in mind, and choose the most possible reason if you think the question cannot be answered based on the text. Output the results in JSON format. \\

\{in\_context\_examples\}\\
Text: \{context\}\\
Question: \{input\}\\
Answer:
\end{minipage}
\end{tabular}
\end{tcolorbox}
\vspace{-2mm}
\caption{Prompt for the failure case analysis.}
\label{tab:prompt_failure}
\end{minipage}
\end{table*}

\begin{table}[h]
\centering
\resizebox{1.0\linewidth}{!}{
\begin{tabular}{l p{16.5cm} }
\toprule
NarrativeQA & You are given a story, which can be either a novel or a movie script, and a question. Answer the question as concisely as you can, using a single phrase if possible. Do not provide any explanation. If the question cannot be answered based on the information in the article, write ``unanswerable''. Story: \{context\} Now, answer the question based on the story as concisely as you can, using a single phrase if possible. Do not provide any explanation. If the question cannot be answered based on the information in the article, write ``unanswerable''. Question: \{input\} Answer: \\ \midrule
Qasper & You are given a scientific article and a question. Answer the question as concisely as you can, using a single phrase or sentence if possible. If the question cannot be answered based on the information in the article, write ``unanswerable''. If the question is a yes/no question, answer ``yes'', ``no'', or ``unanswerable''. Do not provide any explanation. Article: \{context\} Answer the question based on the above article as concisely as you can, using a single phrase or sentence if possible. If the question cannot be answered based on the information in the article, write ``unanswerable''. If the question is a yes/no question, answer ``yes'', ``no'', or ``unanswerable''. Do not provide any explanation. Question: {input} Answer:\\ \midrule
MultiFQA & Read the following text and answer briefly.  \{context\}  Now, answer the following question based on the above text, only give me the answer and do not output any other words. If the question cannot be answered based on the information in the article, write ``unanswerable''.  Question: \{input\} Answer: \\ \midrule
HotpotQA & Answer the question based on the given passages. Only give me the answer and do not output any other words. If the question cannot be answered based on the information in the article, write ``unanswerable''. The following are given passages. \{context\}  Answer the question based on the given passages. Only give me the answer and do not output any other words. If the question cannot be answered based on the information in the article, write ``unanswerable''.  Question: \{input\} Answer:\\ \midrule
2WikiMQA & Answer the question based on the given passages. Only give me the answer and do not output any other words. If the question cannot be answered based on the information in the article, write ``unanswerable''.  The following are given passages. \{context\}  Answer the question based on the given passages. Only give me the answer and do not output any other words. If the question cannot be answered based on the information in the article, write ``unanswerable''.  Question: \{input\} Answer:\\ \midrule
MuSiQue & Answer the question based on the given passages. Only give me the answer and do not output any other words. If the question cannot be answered based on the information in the article, write ``unanswerable''.  The following are given passages. \{context\}  Answer the question based on the given passages. Only give me the answer and do not output any other words. If the question cannot be answered based on the information in the article, write ``unanswerable''.  Question: \{input\} Answer:\\ \midrule
QMSum & You are given a meeting transcript and a query containing a question or instruction. Answer the query in one or more sentences. If the question cannot be answered based on the information in the article, write ``unanswerable''.  Transcript: \{context\}  Now, answer the query based on the above meeting transcript in one or more sentences. If the question cannot be answered based on the information in the article, write ``unanswerable''.  Query: \{input\} Answer:\\ \midrule
EN.QA & Read the book and answer the question. Be very concise in your answer. If the question cannot be answered based on the information in the article, write ``unanswerable''.  \{context\}  Question: \{input\} Only give me the answer and do not output any other words. If the question cannot be answered based on the information in the article, write ``unanswerable''. Answer:\\ \midrule
EN.MC & Read the book and answer the question. If the question cannot be answered based on the information in the article, write ``unanswerable''.  \{context\}  Question: \{input\} \{all\_classes\}  Only output the letter of the correct answer and do not output any other words. If the question cannot be answered based on the information in the article, write ``unanswerable''. The letter of the correct answer is \\
\bottomrule
\end{tabular}
}
\captionof{table}{Prompts for each dataset.}
\label{tab:prompt}  
\end{table}

\end{document}